\documentclass[10pt,twocolumn,letterpaper]{article}

\usepackage{cvpr}
\usepackage{times}
\usepackage{epsfig}
\usepackage{graphicx}
\usepackage{amsmath}
\usepackage{amssymb}

\usepackage{enumitem}
\usepackage{multirow}
\usepackage{makecell}
\usepackage{tabulary}
\usepackage{pifont}
\usepackage{subfigure}

\usepackage[breaklinks=true,bookmarks=false]{hyperref}

\cvprfinalcopy 

\def\cvprPaperID{3807} 

\ifcvprfinal\fi
\begin{document}

\title{Jointly Localizing and Describing Events for Dense Video Captioning\thanks{{\small This work was performed at Microsoft Research Asia.}}}

\author{Yehao Li $^{\dag}$$^{\P}$, Ting Yao $^{\ddag}$, Yingwei Pan $^{\S}$, Hongyang Chao $^{\dag}$$^{\P}$, and Tao Mei $^{\ddag}$ \\
{\small\centering$^{\dag}$ School of Data and Computer Science, Sun Yat-sen University, Guangzhou, China}\\
{\small\centering$^{\ddag}$ Microsoft Research, Beijing, China}\\
{\small\centering$^{\S}$ University of Science and Technology of China, Hefei, China}\\
{\small\centering$^{\P}$ Key Laboratory of Machine Intelligence and Advanced Computing (Sun Yat-sen University), Ministry of Education}\\
{\tt\small \{yehaoli.sysu, panyw.ustc\}@gmail.com, \{tiyao, tmei\}@microsoft.com, isschhy@mail.sysu.edu.cn}
}

\maketitle
\thispagestyle{empty}

\begin{abstract}
Automatically describing a video with natural language is regarded as a fundamental challenge in computer vision. The problem nevertheless is not trivial especially when a video contains multiple events to be worthy of mention, which often happens in real videos. A valid question is how to temporally localize and then describe events, which is known as ``dense video captioning." In this paper, we present a novel framework for dense video captioning that unifies the localization of temporal event proposals and sentence generation of each proposal, by jointly training them in an end-to-end manner. To combine these two worlds, we integrate a new design, namely descriptiveness regression, into a single shot detection structure to infer the descriptive complexity of each detected proposal via sentence generation. This in turn adjusts the temporal locations of each event proposal. Our model differs from existing dense video captioning methods since we propose a joint and global optimization of detection and captioning, and the framework uniquely capitalizes on an attribute-augmented video captioning architecture. Extensive experiments are conducted on ActivityNet Captions dataset and our framework shows clear improvements when compared to the state-of-the-art techniques. More remarkably, we obtain a new record: METEOR of 12.96\% on ActivityNet Captions official test set.
\end{abstract}

\section{Introduction}
\begin{figure}[!tb]
\centering {\includegraphics[width=0.49\textwidth]{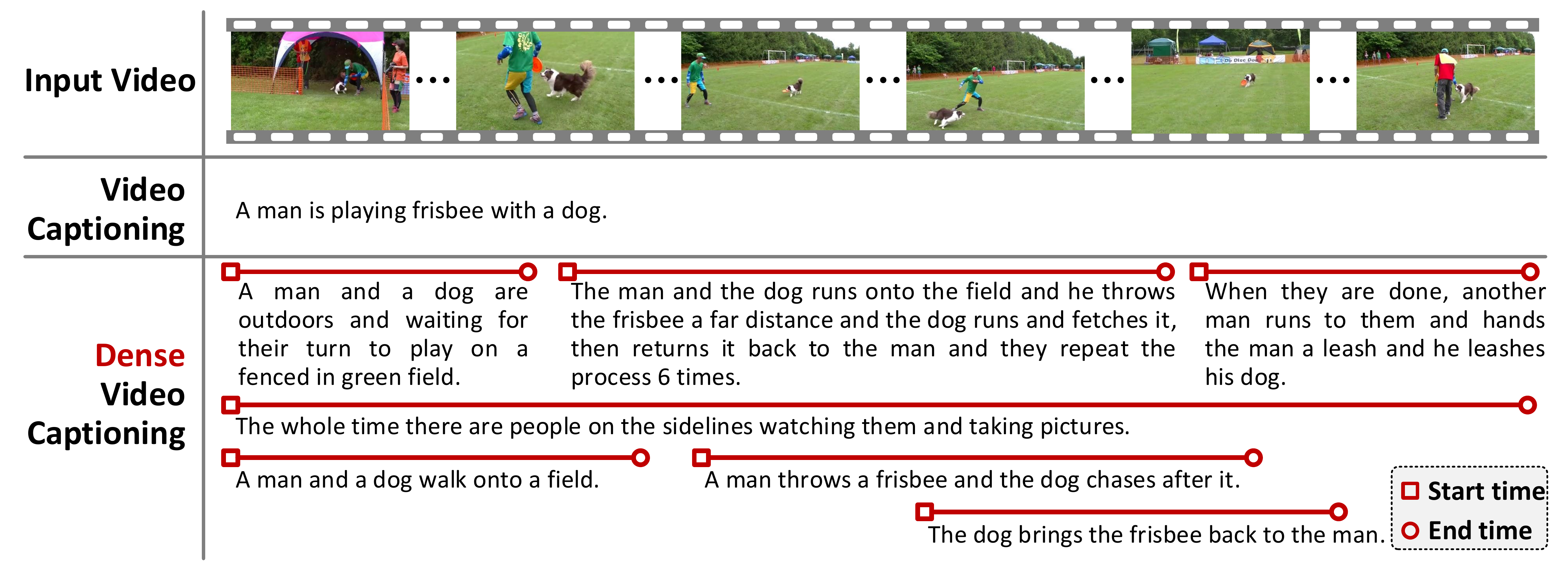}}
\vspace{-0.22in}
\caption{\small Examples of video captioning and dense video captioning (upper row: input video; middle row: the sentence generated by video captioning method; bottom row: temporally localized sentences generated by dense video captioning approach.)}
\label{fig:figintro}
\vspace{-0.23in}
\end{figure}

The recent advances in 2D and 3D Convolutional Neural Networks (CNNs) have successfully pushed the limits and improved the state-of-the-art of video understanding. For instance, the first rank performance achieves 8.8\% in terms of top-1 error in untrimmed video classification task of ActivityNet Challenge 2017 \cite{ActivityNet}. As such, it has become possible to recognize a video with a pre-defined set of labels or categories. In a further step to describe a video with a complete and nature sentence, video captioning \cite{Pan:CVPR16,Venugopalan:ICCV15,Yao:ICCV15} has expanded the understanding from individual labels to a sequence of words to express richer semantics and relationships in the video. Nevertheless, considering that videos in real life are usually long and contain multiple events, the conventional video captioning methods generating only one caption for a video in general will fail to recapitulate all the events in the video. Take the video in Figure \ref{fig:figintro} as an example, the output sentence generated by a popular video captioning method \cite{Venugopalan:ICCV15} is unable to describe the procedure of ``playing frisbee with a dog" in detail. As a result, the task of dense video captioning is introduced recently in \cite{krishna2017dense} and the ultimate goal is to generate a sentence for each event occurring in the video.

The difficulty of dense video captioning originates from two aspects: 1) how to accurately localize each event in time? 2) how to design a powerful sentence generation model? In the literature, there have been several techniques, including temporal action proposal \cite{buchsst,caba2016fast,escorcia2016daps,lin2017single} and image/video captioning \cite{Venugopalan:ICCV15,Yao:ICCV15,yao2017boosting}, being proposed for each individual aspect. However, simply solving the problem of dense video captioning in a two-stage way, i.e., first temporal event proposal and then sentence generation, may destroy the interaction between localizing and describing events, resulting in a sub-optimal solution.

This paper proposes a novel deep architecture to unify the accurate localization of temporal events with the descriptive principle of sentence generation for dense video captioning. Technically, we devise a new descriptiveness regression component and integrate it into a single shot detection framework as a bridge, on one hand to measure the complexity of each event being described in sentence generation, and on the other, to adjust the event proposal. More specifically, the descriptiveness regression guides the learning of temporal event proposal together with event/background classification and temporal coordinate regression. In between, event/background classification is to predict event proposals and temporal coordinate regression is to refine the temporal boundaries of each proposal. Furthermore, the inference of descriptiveness regression is employed as an ``attention" to weight video clips in each proposal locally. The proposal-level representation is then averaged over all the clip-level representations in the proposal weighted by the holistic attention score of each clip and finally fed into an attribute-augmented captioning architecture for sentence generation. As such, the task of dense video captioning could be jointly learnt and globally optimized in an end-to-end manner.

The main contribution of this work is the proposal of a new architecture to unify the temporal localization of event proposals and sentence generation for dense video captioning. The solution also leads to the elegant views of what kind of interaction should be built across the two sub problems and how to model and integrate the interaction in a deep learning framework, which are problems not yet fully understood in the literature.

\section{Related Work}\label{sec:RW}
\textbf{Temporal Action Proposal.}
\cite{duchenne2009automatic} is one of the early works that detects temporal segments containing the action of interest in a sliding windows fashion. Next, a few subsequent works \cite{buchsst,caba2016fast,escorcia2016daps,gao2017turn,shou2016temporal} tackle temporal action proposal by leveraging action classifiers on a smaller number of temporal windows. In particular, Sparse-prop \cite{caba2016fast} utilizes dictionary learning to encode representations of trimmed action instances and then retrieves the most representative segments from testing videos, which are treated as the class-independent proposals. S-CNN \cite{shou2016temporal} trains a 3D CNNs to classify a video segment as background or being-action and employs varied length temporal windows for multi-scale action proposal generation. Later in \cite{escorcia2016daps}, DAPs utilizes Long Short-Term Memory (LSTM) to encode video streams and enables multi-scale proposal generation inside the streams with a single pass through the video, obviating the need for deploying sliding windows on multiple scales. Furthermore, Buch \emph{et al.} develop SST based on DAPs by constructing no overlapping sliding windows over the input video and encoding each window sequentially with a Gated Recurrent Unit (GRU) in \cite{buchsst}. Most recently, Gao \emph{et al.} \cite{gao2017turn} design temporal coordinate regression for temporal action proposal generation.

\textbf{Video Captioning.}
The research in this direction has proceeded along two different dimensions: template-based language methods \cite{Guadarrama:ICCV13,Kojima:IJCV02,Rohrbach:ICCV13} and sequence learning approaches (e.g., RNNs) \cite{krishna2017dense,Pan:CVPR16,Venugopalan:ICCV15,Venugopalan:NAACL15,Yao:ICCV15,Yu:CVPR16}. Template-based language methods directly generate the sentence with detected keywords in predefined language template. Sequence learning approaches utilize CNN plus RNN architecture to generate novel sentences with more flexible syntactical structures. In \cite{Venugopalan:NAACL15}, Venugopalan \emph{et al.} present a LSTM-based model to generate video descriptions with the mean pooling representation over all frames. The framework is then extended by inputting both frames and optical flow images into an encoder-decoder LSTM in \cite{Venugopalan:ICCV15}. Compared to mean pooling, Yao \emph{et al.} propose to utilize the temporal attention mechanism to exploit temporal structure for video captioning \cite{Yao:ICCV15}. Later in \cite{Yu:CVPR16}, a hierarchical RNN is devised to further capture the inter-sentence dependency, targeting for describing a long video with a paragraph consisting of multiple sentences. Different from the video paragraph captioning with non-overlapping and annotated temporal intervals, a more challenge task, named as dense video captioning, is recently introduced in \cite{krishna2017dense} which involves both detecting and describing multiple events in a video. A two-stage dense-captioning system is thus designed by leveraging DAPs \cite{escorcia2016daps} to localize temporal event proposals and a LSTM-based sequence learning module to describe each event proposal. Most recently, \cite{yao2017msr} additionally incorporates KNN-based retrieval module into LSTM-based sequence learning module to boost video captioning.

\textbf{Summary.} Our work aims to detect and describe events in video, i.e., dense video captioning. Different from the aforementioned method \cite{krishna2017dense}, our approach contributes by studying not only detecting the events with the simple objective of binary classification (i.e., event or background) and modeling sentence generation with LSTM, but also enhancing the temporal event proposal by utilizing both temporal boundary regression to correct start and end time of event and descriptiveness regression to infer whether the event can be well described from language perspective. Moreover, sentence generation module is further boosted by leveraging semantic attributes and reinforcement learning to optimize LSTM with non-differentiable metrics.

\begin{figure*}[!tb]
\centering {\includegraphics[width=0.9\textwidth]{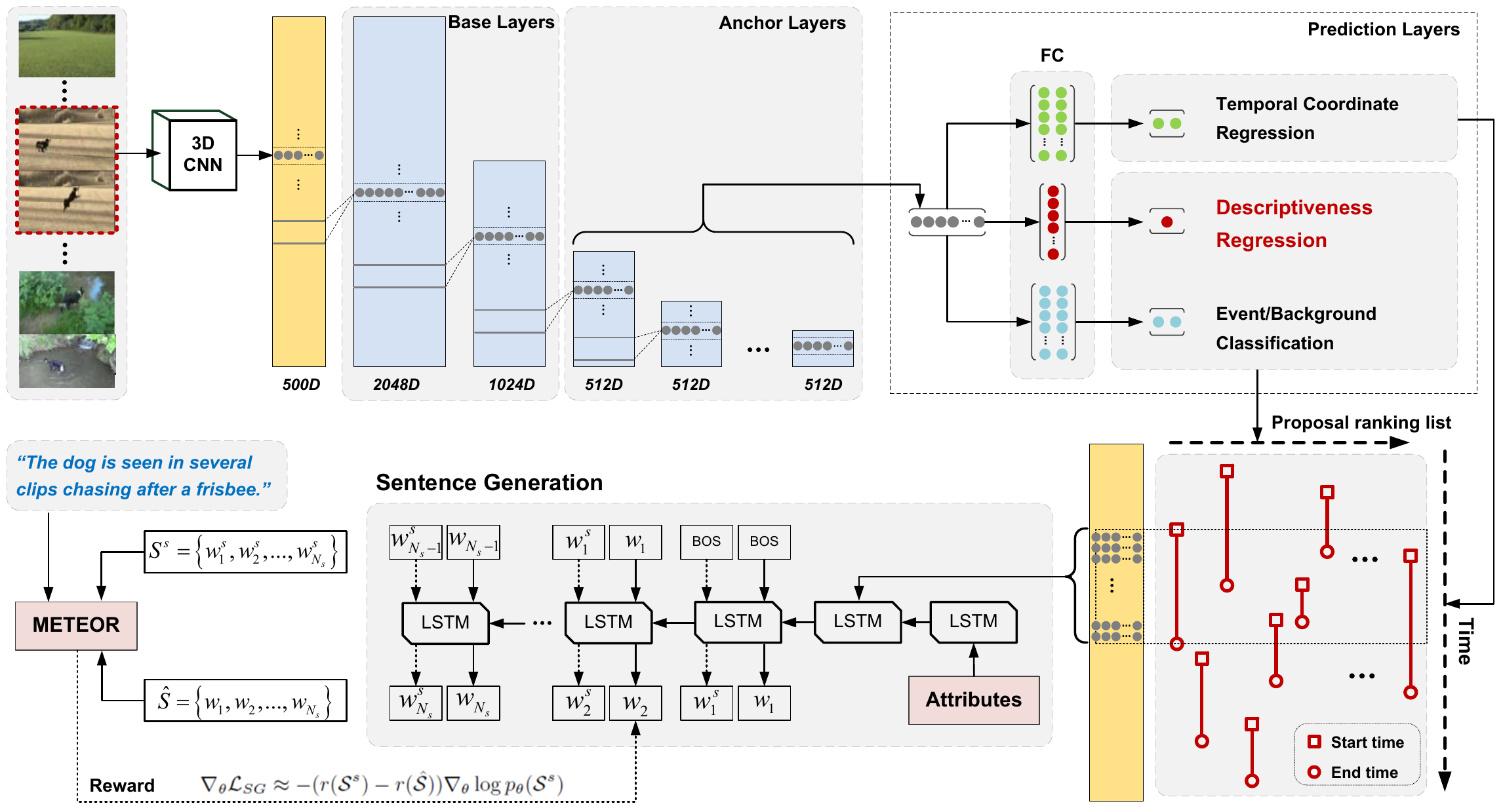}}
\vspace{-0.05in}
\caption{\small An overview of our Dense Video Captioning framework mainly including Temporal Event Proposal (TEP) and Sentence Generation (SG) (better viewed in color). The input video is first encoded into a series of clip-level features via a 3-D CNNs, which are fed into TEP module to produce candidate proposals. The TEP module is employed by integrating the event/background classification to predict event proposal, temporal coordinate regression to refine the temporal boundaries of each proposal, and descriptiveness regression to infer the descriptive complexity of each event, into a single shot detection architecture. After ranking the candidate proposals with regard to both eventness and descriptiveness scores, the top proposals are in turn injected into SG module for sentence generation. The SG module leverages both attributes and reinforcement learning based optimization to enhance captioning.}
\label{fig:fig1}
\vspace{-0.2in}
\end{figure*}

\section{Dense Video Captioning}
The basic idea of this work is to automatic describe multiple events in videos by temporally localizing event proposals and generating language sentence for each event proposal. The temporal event proposal (TEP) is performed by encapsulating the event classification to recognize video segments of events from backgrounds, proposal generation to temporally localize the event, and descriptiveness inference procedure to infer the descriptive complexity of this event, in one single network. Such design enables straightforward temporal event proposal in a single shot manner to ease the training consumption. The sentence generation module (SG) leverages an attribute-augmented LSTM-based model for generating descriptions. Moreover, the policy-gradient based reinforcement learning is adopted to optimize LSTM with evaluation metric based reward, harmonizing the module with respect to the testing inference. Please note that the descriptiveness inference procedure in TEP module is not only leveraged to additionally refine the localized event proposal from language perspective through descriptiveness regression, but also integrated into SG module to consider the descriptiveness scores as one kind of temporal attention over clips for weighted fusing them as the input proposal-level representation of LSTM. As such, our system including both TEP and SG modules can be jointly trained through the global optimization of detection and captioning in an end-to-end manner. An overview of our dense video captioning system is illustrated in Figure \ref{fig:fig1}.

\subsection{Problem Formulation}
Suppose we have a video $\mathcal{V}=\{v_t\}^{T_v}_{t=1}$ with $T_v$ frames/clips and $v_t$ denotes the $t$-th frame/clip in temporal order. The ultimate target of our dense video captioning system is to generate a set of temporal localized descriptions ${\Phi _v} = \{ {\phi _i} = (t^i_{start},t^i_{end},{\mathcal{S}_i})\} _{i = 1}^{{M_{v}}}$ for the input video $\mathcal{V}$, where $M_{v}$ is the number of sentences, $t^i_{start}$ and $t^i_{end}$ represent the starting time and ending time for each sentence $\mathcal{S}_i$, and ${\mathcal{S}_i= \{w_1, w_2, ..., w_{N_s}\}}$ consists of $N_s$ words.

Hence the TEP module in our system is firstly utilized to produce a set of candidate proposals for the input video $\mathcal{V}$:
\vspace{-0.05in}
\begin{equation}\label{Eq:EqPF1}\small
{\Phi _p} = \{ {\phi^i_{{p}}}=(t^i_{start}, t^i_{end}, p_{event}^i, p_{des}^i)\} _{i = 1}^{{N_p}},
\vspace{-0.05in}
\end{equation}
where $p_{event}^i$ is the probability of recognizing the candidate as an event (i.e., eventness score), $p_{des}^i$ denotes the descriptiveness score measuring how well the candidate can be described from language perspective and $N_p$ is the total number of candidate proposals. By consolidating the idea of selecting proposals from both vision and language perspectives, all the candidates are ranked according to the fused score $p_{conf}^i = p_{event}^i + {\lambda_0} p_{des}^i$ and only the candidates with a $p_{conf}^i$ higher than a threshold are injected into SG module for captioning, denoted as ${\Phi _{\hat p}}$.

Inspired by the successes of sequence learning models in machine translation \cite{Sutskever:NIPS14} and attributes utilized in image/video captioning \cite{Fang:CVPR15,pan2017video,yao2017incorporating}, we formulate our SG module as an attribute-argument LSTM-based model which encodes the input event proposal representation (${\bf{F}}$) and its detected attributes/categories (${\bf{A}}$) into a fixed dimensional vector and then decodes it to the output target sentence. As such, the sentence generation problem we exploit here can be formulated by minimizing the negative $\log$ probability of the correct textual sentence ($ -\log {\Pr{({\mathcal {S}}|{\bf{F}}, {\bf{A}})}}$). The negative $\log$ probability is typically measured with cross entropy loss, resulting in the discrepancy of evaluation between training and inference. Hence, to further boost our SG module by amending such discrepancy, we take the inspiration from the reinforcement learning \cite{ranzato2015sequence} leveraged in sequence learning and directly optimize LSTM by minimizing the following expected sentence-level reward loss as
\vspace{-0.05in}
\begin{equation}\label{Eq:EqPF2}\small
{\mathcal{L}_{SG}} =  - {{\mathbb E}_{\mathcal{S} \sim {p_\theta }}}[r(\mathcal{S})],
\vspace{-0.05in}
\end{equation}
where $\theta$ denotes the parameters of LSTM that schedule a policy ${p_\theta }$ for generating sentence. $r(\mathcal{S})$ is the reward measured by comparing the generated sentence $\mathcal{S}$ to ground-truth sentences over non-differentiable evaluation metric.

\subsection{Temporal Event Proposal}
Existing solutions for temporal action/event proposal mainly focus on detecting events with binary classifier (i.e., event or background) in a sliding windows fashion. However, the temporal sliding windows are typically too dense and even designed with multiple scales, resulting in the heavy computation cost. Inspired from spatial boundary regression in object localization \cite{ren2015faster} which simultaneously predicts objectness score and object bound, we integrate the event classification with temporal coordinate regression for correcting event's temporal bound, pursuing both low-cost and high-quality event proposals. Moreover, for the dense video captioning task, the event identification undoubtedly plays the major role in temporal event proposal, while the proposals containing rich describable objects or scenes are also preferred by human beings in description generation. As such, a novel descriptiveness regression is especially devised to infer the descriptive complexity of each proposal, and further refine the event proposal from language perspective. Similar to single shot object detection in \cite{liu2016ssd}, all the three components (i.e, event classification, temporal coordinate regression and descriptiveness regression) are elegantly integrated into one feed-forward CNNs, aiming to simultaneously produce a fixed-size set of proposals, scores for the presence of event in the proposals (i.e., eventness scores), and descriptiveness scores of proposals.

\textbf{Input.} Technically, given input video $\mathcal{V}=\{v_t\}^{T_v}_{t=1}$, a 3-D CNN is utilized to encode the frame sequence into a series of clip-level features $\{ {{\bf{f}}_t} = F({v_t}:{v_{t + \delta }})\}^{T_f}_{t=1}$ where $\delta$ is the temporal resolution of each feature ${{\bf{f}}_t}$. In our experiments, C3D \cite{tran2015learning} is adopted as 3-D CNNs encoder $F$ with $\delta=16$ frames and the temporal interval for encoding is set as $8$ frames, resulting in the initial output feature map with the size of ${T_f} \times D_0$. Note that $D_0$ is the dimension of clip-level feature and $T_f = T_v / 8$ discretizes the video frames.

\textbf{Network Architecture.} The initial feature map of size ${T_f} \times D_0$ is fed into 1-D CNNs architecture in TEP module, which consists of convolutional layers in three groups: base layers, anchor layers, and prediction layers, as shown in Figure \ref{fig:fig1}. In particular, two base layers ($conv_1$ and $conv_2$) are firstly devised to reduce the temporal dimension of feature map and increase the size of temporal receptive fields, producing the output feature map of size ${T_f/2} \times 1024$. Then, we stack nine anchor layers ($conv_3$ to $conv_{11}$) on the top of base layer $conv_2$, each of which is designed with the same configuration (kernel size: 3, stride size: 2, and filter number: 512). These anchor layers decrease in temporal dimension of feature map progressively, enabling the temporal event proposals at multiple temporal scales.

For each anchor layer, its output feature map is injected into prediction layer to produce a fixed set of predictions in one shot manner. Concretely, given an output feature map $f_j$ with the size of ${T_{f_j}} \times D_j$, the basic element (anchor) for predicting parameters of a candidate proposal is a $1 \times D_j$ feature map cell that produces a prediction score vector ${p_{pred}} = ({p_{cls}},\Delta c,\Delta w,{p_{des}})$ via fully connected layers. $p_{cls}=[p_{event}, p_{bk}]$ denotes the two dimensional classification scores for event/background and ${p_{des}}$ is the descriptiveness score to infer the confidence of this proposal to be well described. $\Delta c$ and $\Delta w$ are two temporal offsets relative to the default center location ${\mu _c}$ and width ${\mu _w}$ of this anchor, which are leveraged to adjust its temporal coordinates as
\begin{equation}\small
\begin{array}{l}
{\varphi _c} = {\mu _c} + {\alpha _1}{\mu _w}\Delta c,~~~{\varphi _w} = {\mu _w} \exp ({\alpha _2} \Delta w),\\
t_{start} = {\varphi _c} - \frac{1}{2} {\varphi _w},~~~t_{end} = {\varphi _c} + \frac{1}{2} {\varphi _w},
\end{array}
\end{equation}
where ${\varphi _c}$ and ${\varphi _w}$ are refined center location and width of the anchor. $\alpha _1$ and $\alpha _2$ are utilized to control the impact of temporal offsets, both of which are set as $0.1$. $t_{start}$ and $t_{end}$ represent the adjusted starting and ending time of the anchor. In addition, derived from the anchor boxes in \cite{liu2016ssd,ren2015faster}, we associate a set of default temporal boundaries with each feature map cell. The different temporal scale ratios for these default temporal boundaries are denoted as ${R_s} = \{ {r_d}\} _{d = 1}^{{D_s}}=\{1, 1.25, 1.5\}$. For each temporal scale ratio $r_d$, we can thus achieve one default center location (${\mu _{c_d}}=\frac{{t+0.5}}{{T_{f_j}}}$) and width (${\mu _{w_d}}=\frac{{r_d}}{{T_{f_j}}}$) of $t$-th feature map cell, resulting in a total of ${T_{f_j}}{D_s}$ anchors. Accordingly, for the feature map $f_j$, the set of all the produced proposals is defined as ${\Phi _{f_j}} = \{ {\phi _{f_j}} = (t_{start}^u,t_{end}^u,p_{cls}^u,p_{des}^u)\} _{u = 1}^{{N_U}}$, where $N_U = {T_{f_j}}{D_s}$. By accumulating all the produced proposals of the output feature maps of nine anchor layers, the final predicted proposal set is ${\Phi _p} = \{ \Phi _{f_j}\}=\{ \phi^{i}_{p}\}^{N_p}_{i=1}$.

\textbf{Training.} During training, a positive/negative label should be firstly assigned to each predicted proposal conditioned on the ground truth proposal set ${\Phi _v}$. Specifically, for each $\phi^{i} _{p} \in {\Phi _p}$, we measure its temporal Intersection over Union (tIoU) with each ground truth proposal and obtain the highest tIoU. If the highest tIoU is larger than 0.7, $\phi^{i} _{p}$ is treated as positive sample with regard to the corresponding ground truth proposal $\phi _{g}$, otherwise $\phi^{i} _{p}$ is a negative sample. The training objective in our TEP module is formulated as a multi-task loss by integrating the event/background classification loss ($\mathcal{L}_{event}$) for distinguishing events from backgrounds, temporal coordinate regression loss ($\mathcal{L}_{tcr}$) for adjusting temporal coordinate of event proposal, and descriptiveness regression loss ($\mathcal{L}_{des}$) for inferring the descriptive complexity of each proposal, which is defined as
\vspace{-0.05in}
\begin{equation}\label{Eq:EqTEP1}\small
\mathcal{L}_{TEP} = {\mathcal{L}_{event}} + \alpha {\mathcal{L}_{tcr}} + \beta {\mathcal{L}_{des}},
\vspace{-0.05in}
\end{equation}
where $\alpha$ and $\beta$ are tradeoff parameters. The event/background classification loss $\mathcal{L}_{event}$ is measured as the standard softmax loss over the whole predicted proposal set ${\Phi _p}$. The temporal coordinate regression loss $\mathcal{L}_{tcr}$ is devised as Smooth L1 loss \cite{girshick2015fast} ($S_{L1}$) between the positive predicted proposals and the corresponding ground truth proposals. Similar to \cite{ren2015faster}, both of the temporal offsets for center location (${\varphi _c}$) of the predicted proposal and for its width (${\varphi _w}$) are regressed as
\vspace{-0.05in}
\begin{equation}\small
{\mathcal{L}_{tcr}} = \frac{1}{{{N_{pos}}}}\sum\limits_{i = 1}^{{N_{p}}} {\mathbb I}_{i} {(S_{L1}({\varphi ^{i}_{c}} - {g^{i}_{c}}) + S_{L1}({\varphi^{i} _{w}} - {g^{i}_{w}}))},
\vspace{-0.05in}
\end{equation}
where ${\mathbb I}_{i}$ denotes the assigned label for predicted proposal $\phi^{i}_{p}$ (1 for positive sample and 0 for negative sample), $N_{pos}$ is the number of positive samples, $g^{i}_{c}$ and $g^{i}_{w}$ represent the center location and width of ground truth proposal.

The descriptiveness regression loss $\mathcal{L}_{des}$ is calculated as the Euclidean distance between the inferred descriptiveness score $p^{i}_{des}$ of the predicted proposal and its sentence-level reward $r({\mathcal{S}_i})$ in SG with regard to ground truth sentence:
\vspace{-0.05in}
\begin{equation}\small
{\mathcal {L}_{des}} = \frac{1}{{{N_{p}}}}\sum\limits_{i = 1}^{{N_{p}}} {\left\| {p_{des}^i - r({\mathcal{S}_i})} \right\|_2^2}.
\vspace{-0.05in}
\end{equation}
In particular, for each positive predicted proposal ${\phi^i _p}$, we achieve its sentence-level reward $r({\mathcal{S}_i})$ by directly feeding this proposal into SG module and comparing the generated sentence ${\mathcal{S}_i}$ with the corresponding ground-truth sentence over evaluation metric (e.g., METEOR). For each negative sample, its sentence-level reward is naturally fixed as 0. Accordingly, by minimizing the descriptiveness regression loss, our prediction layer is additionally endowed with the ability to directly infer the approximate descriptiveness score (i.e., the sentence-level reward for captioning) of an event proposal without referring ground-truth sentences.

\subsection{Sentence Generation}
Given the set of selected predicted event proposals ${\Phi _{\hat p}} \subset {\Phi _{p}}$ from TEP module, each proposal ${\phi _{\hat p}} \in {\Phi _{\hat p}}$ is injected into attribute-augmented LSTM-based model in SG module for description generation. Specifically, the attributes representation $\bf{A}$ of predicted proposal ${\phi _{\hat p}}$ is firstly transformed into LSTM to inform the whole LSTM about the high-level attributes, followed by the proposal representation $\bf{F}$ which is encoded into LSTM at the second time step. Then, LSTM decodes each output word based on previous word and previous step's LSTM hidden state.

\textbf{Descriptiveness-driven Temporal Attention.} One natural way to achieve the proposal representation $\bf{F}$ is performing ``mean pooling" process over all the clips within this proposal. However, in many cases, the generated description only relates to some key clips with low descriptive complexity. As a result, to pinpoint the local clips containing rich describable objects or scenes and further incorporate the contributions of different clips into producing proposal representation, a descriptiveness-driven temporal attention mechanism is employed on the predicted event proposal. Given the input proposal ${\phi _{\hat p}}$ containing ${N_{\hat p_{i}}}$ clips $\{v_i\}^{N_{\hat p_{i}}}_{i=1}$, the clip-level descriptiveness score $p_{_{des}}^{{v_i}}$ of each clip $v_i$ is firstly achieved by holistically taking the average of all the descriptiveness scores of predicted proposals containing this clip $v_i$. Here we treat the clip-level descriptiveness score of each clip as one kind of temporal attention over all the clips within this proposals. Based on the attention distribution, we calculate the weighted sum of the clip features and obtain the aggregated proposal feature ${\bf{F}}$ weighted by holistic attention score of each clip:
\vspace{-0.05in}
\begin{equation}\small
{\alpha _{{v_i}}} = {{p_{des}^{{v_i}}} \mathord{\left/
 {\vphantom {{p_{des}^{{v_i}}} {\sum\limits_{j = 1}^{{N_{{\hat p_i}}}} {p_{des}^{{v_j}}} }}} \right.
 \kern-\nulldelimiterspace} {\sum\limits_{j = 1}^{{N_{{\hat p_i}}}} {p_{des}^{{v_j}}} }},~~~{{\bf{F}}} = \sum\limits_{i = 1}^{{N_{\hat p_{i}}}} {{\alpha _{{v_i}}} \cdot {{\bf{f}}_{_{{v_i}}}}},
\vspace{-0.05in}
\end{equation}
where $\alpha _{v_i}$ is the normalized attention score and ${{\bf{f}}_{_{{v_i}}}}$ is the clip representation of $v_i$. The aggregated proposal feature could be regarded as a more informative proposal feature since the most descriptive clips for sentence generation have been distilled with higher attention weights.

\textbf{Training.} The training objective in our SG module is formulated as the expected sentence-level reward loss in Eq.(\ref{Eq:EqPF2}). Inspired from Self-critical Sequence Training (SCST) \cite{Rennie:CVPR17}, the gradient of this objective is given by
\vspace{-0.05in}
\begin{equation}\small
{\nabla _\theta } {\mathcal {L}_{SG}} \approx  - (r({\mathcal{S}^s}) - r(\mathcal{\hat S})){\nabla _\theta }\log {p_\theta }({\mathcal{S}^s}) ,
\vspace{-0.05in}
\end{equation}
where $\mathcal{S}^s$ is a sampled sentence and $r(\mathcal{\hat S})$ denotes the reward of baseline achieved by greedily decoding inference.

\subsection{Joint Detection and Captioning}
The overall objective of our dense video captioning is comprised of the training objective of TEP module in Eq.(\ref{Eq:EqTEP1}) and the reward loss of SG module in Eq.(\ref{Eq:EqPF2}):
\vspace{-0.05in}
\begin{equation}\label{Eq:EqJDC}\small
\mathcal{L} = \lambda_1 \mathcal{L}_{TEP} + \lambda_2 \mathcal{L}_{SG},
\vspace{-0.05in}
\end{equation}
where $\lambda_1$ and $\lambda_2$ are tradeoff parameters for TEP and SG, respectively. Note that descriptiveness inference could be regarded as a bridge which is not only leveraged in TEP for adjusting the event proposal from language perspective, but also integrated into SG for measuring the descriptiveness-driven temporal attention to boost sentence generation, enabling the interaction between TEP and SG. As a result, the overall objective function of our system can be solved through the joint and global optimization of detection and captioning in an end-to-end manner.

\section{Experiments}
We conduct our experiments on the ActivityNet Captions dataset \cite{krishna2017dense} and evaluate our proposed system on both dense video captioning and temporal event proposal tasks.

\subsection{Dataset}
The dataset, ActivityNet Captions, is a recently collected large-scale dense video captioning benchmark, which contains 20,000 videos covering a wide range of complex human activities. Each video is aligned with a series of temporally annotated sentences. On average, there are 3.65 temporally localized sentences for each video, resulting in a total of 100,000 sentences. In our experiments, we follow the settings in \cite{krishna2017dense}, and take 10,024 videos for training, 4,926 for validation and 5,044 for testing.

\begin{table*}[t]\scriptsize
    \centering
    \caption{\small BLEU@$N$ (B@$N$), METEOR (M) and CIDEr-D (C) scores of our DVC and other state-of-the-art video captioning methods for dense video captioning task on ActivityNet Captions validation set (left: performances with ground truth (GT) proposals; right: performances with the learnt proposals from our TEP module). All values are reported as percentage (\%).}
    \begin{tabular}{l | c c c c c c | c c c c c c}
        \Xhline{2\arrayrulewidth}
		  & \multicolumn{6}{c|}{\textbf{with GT proposals}} & \multicolumn{6}{c}{\textbf{with learnt proposals}} \\
		               & B@1            & B@2           & B@3           & B@4           & M              & C              & B@1            & B@2           & B@3           & B@4           & M             & C              \\
		\hline \hline
		~LSTM~~\cite{Venugopalan:NAACL15}    & 18.40          & 8.76          & 3.99          & 1.53          & 8.66           & 24.07          & 11.19          & 4.73          & 1.75          & 0.60          & 5.53          & 12.06          \\
		~S2VT~~\cite{Venugopalan:ICCV15}     & 18.25          & 8.68          & 4.02          & 1.57          & 8.74           & 24.05          & 11.10          & 4.68          & 1.83          & 0.65          & 5.56          & 12.16          \\
		~TA~~\cite{Yao:ICCV15}               & 18.19          & 8.62          & 3.98          & 1.56          & 8.75           & 24.14          & 11.06          & 4.66          & 1.78          & 0.65          & 5.62          & 12.19          \\
		~H-RNN~~\cite{Yu:CVPR16} & 18.41          & 8.80          & 4.08          & 1.59          & 8.81           & 24.17          & 11.21          & 4.79          & 1.90          & 0.70          & 5.68          & 12.35          \\
        ~DCE~~\cite{krishna2017dense} & 18.13          & 8.43          & 4.09          & 1.60          & 8.88           & 25.12          & 10.81          & 4.57          & 1.90          & 0.71          & 5.69          & 12.43          \\
		\hline
		~DVC-D~~                       & 18.25          & 8.56          & 4.15          & 1.63          & 8.94           & 25.46          & 10.83          & 4.61          & 1.91          & 0.73          & 5.79          & 12.86          \\
		~DVC-D-A~~                     & 19.34          & 9.55          & 4.51          & \textbf{1.71} & 9.31           & \textbf{26.26} & 11.78          & 5.24          & 2.05          & \textbf{0.74} & 6.14          & \textbf{13.21} \\
		~DVC~~                         & \textbf{19.57} & \textbf{9.90} & \textbf{4.55} & 1.62          & \textbf{10.33} & 25.24          & \textbf{12.22} & \textbf{5.72} & \textbf{2.27} & 0.73          & \textbf{6.93} & 12.61          \\
		\Xhline{2\arrayrulewidth}
    \end{tabular}
	\vspace{-0.23in}
    \label{tab:densecaption}
\end{table*}

\subsection{Dense Video Captioning Task}
We firstly investigate our system on dense video captioning task. The task is to detect individual events and then describe each event with natural language.

\textbf{Compared Approaches.}
To empirically verify the merit of our proposed dense video captioning system, we compare the following video captioning baselines:

(1) Long Short-Term Memory (LSTM) \cite{Venugopalan:NAACL15}: LSTM utilizes a CNN plus RNN framework to directly translate from video pixels to natural language descriptions. The frame features are mean pooled to generate the video features.

(2) Sequence to Sequence - Video to Text (S2VT) \cite{Venugopalan:ICCV15}: S2VT incorporates both RGB and optical flow inputs, and the encoding and decoding of the inputs and word representations are learnt jointly in a parallel manner.

(3) Temporal Attention (TA) \cite{Yao:ICCV15}: TA combines the frame representation from GoogleNet \cite{szegedy2015going} and video clip representation from 3D CNN trained on hand-crafted descriptors. Furthermore, a soft attention mechanism is employed to dynamically attend to specific temporal regions of the video while generating sentences.

(4) Hierarchical Recurrent Neural Networks (H-RNN) \cite{Yu:CVPR16}: H-RNN generates paragraphs by using one RNN to generate individual sentence and the second to capture the inter-sentence dependencies. Moreover, both spatial and temporal attention mechanisms are leveraged in H-RNN.

(5) Dense-Captioning Events (DCE) \cite{krishna2017dense}: DCE leverages a multi-scale variant of DAPs \cite{escorcia2016daps} to localize temporal event proposals and S2VT \cite{Venugopalan:ICCV15} as base captioning module to describe each event. An attention module is further incorporated to exploit temporal context for dense captioning.

(6) Dense Video Captioning (DVC) is our complete system in this paper. Two slightly different settings of DVC are named as DVC-D and DVC-D-A. The former only incorporates the descriptiveness-driven temporal attention mechanism into LSTM in SG module and is trained without attributes and reinforcement learning based optimization, while the latter is more similar to DVC that only replaces the expected sentence-level reward loss in DVC with the traditional cross entropy loss.

Note that DCE is the only existing work on dense video captioning task and most previous video captioning works (e.g., LSTM, S2VT, TA, and H-RNN) focus on describing entire videos without detecting a series of events. Hence we compare the five video captioning baselines on dense video captioning task by feeding them with the fixed ground truth proposals or the learnt ones from our TEP module.

\textbf{Settings.}
For video clip representation, we utilize the publicly available 500-way C3D in \cite{krishna2017dense}, whose dimension is reduced by PCA from the original 4,096-way output of $fc7$ of C3D pre-trained on Sports-1M video dataset \cite{karpathy2014large}. For representation of attributes/categories, we treat all the 200 categories on Activitynet dataset \cite{caba2015activitynet} as the high-level semantic attributes and train the attribute detectors with cross entropy loss, resulting in the final 200-way vector of probabilities. Each word in the sentence is represented as ``one-hot" vector (binary index vector in a vocabulary). The dimension of the input and hidden layers in LSTM are both set to 1,024. The tradeoff parameter $\lambda_0$ leveraging the event probability and descriptiveness score for proposal selection is empirically set to 0.2. The tradeoff parameters $\alpha$ and $\beta$ in Eq.(\ref{Eq:EqTEP1}) are set as 0.5 and 10. For the tradeoff parameters $\lambda_1$ and $\lambda_2$ in Eq.(\ref{Eq:EqJDC}), we set them as 1 and 20, respectively.

We mainly implement our DVC based on Caffe \cite{Jia:MM14}, which is one of widely adopted deep learning frameworks. The whole system is trained by Adam \cite{kingma2014adam} optimizer. The initial learning rate is set as 0.00001 and the mini-batch size is set as 1. Note that SG module in our DVC is pre-trained with ground-truth proposal-sentence pairs. The sentence-level reward in SG is measured with METEOR.

\textbf{Evaluation Metrics.}
For the evaluation of our proposed models, we follow the metrics in \cite{krishna2017dense} to measure the ability to jointly localize and describe dense events. This metric computes the mean average precision (mAP) across tIoU thresholds of 0.3, 0.5, 0.7, and 0.9 when captioning the top 1,000 proposals. The precision of captions is measured by three evaluation metrics: BLEU@$N$ \cite{Papineni:ACL02}, METEOR \cite{Banerjee:ACL05}, and CIDEr-D \cite{vedantam2015cider}. All the metrics are computed by using the codes\footnote{\url {https://github.com/ranjaykrishna/densevid_eval}} released by ActivityNet Evaluation Server.

\begin{figure}[!tb]
   \centering {\includegraphics[width=0.46\textwidth]{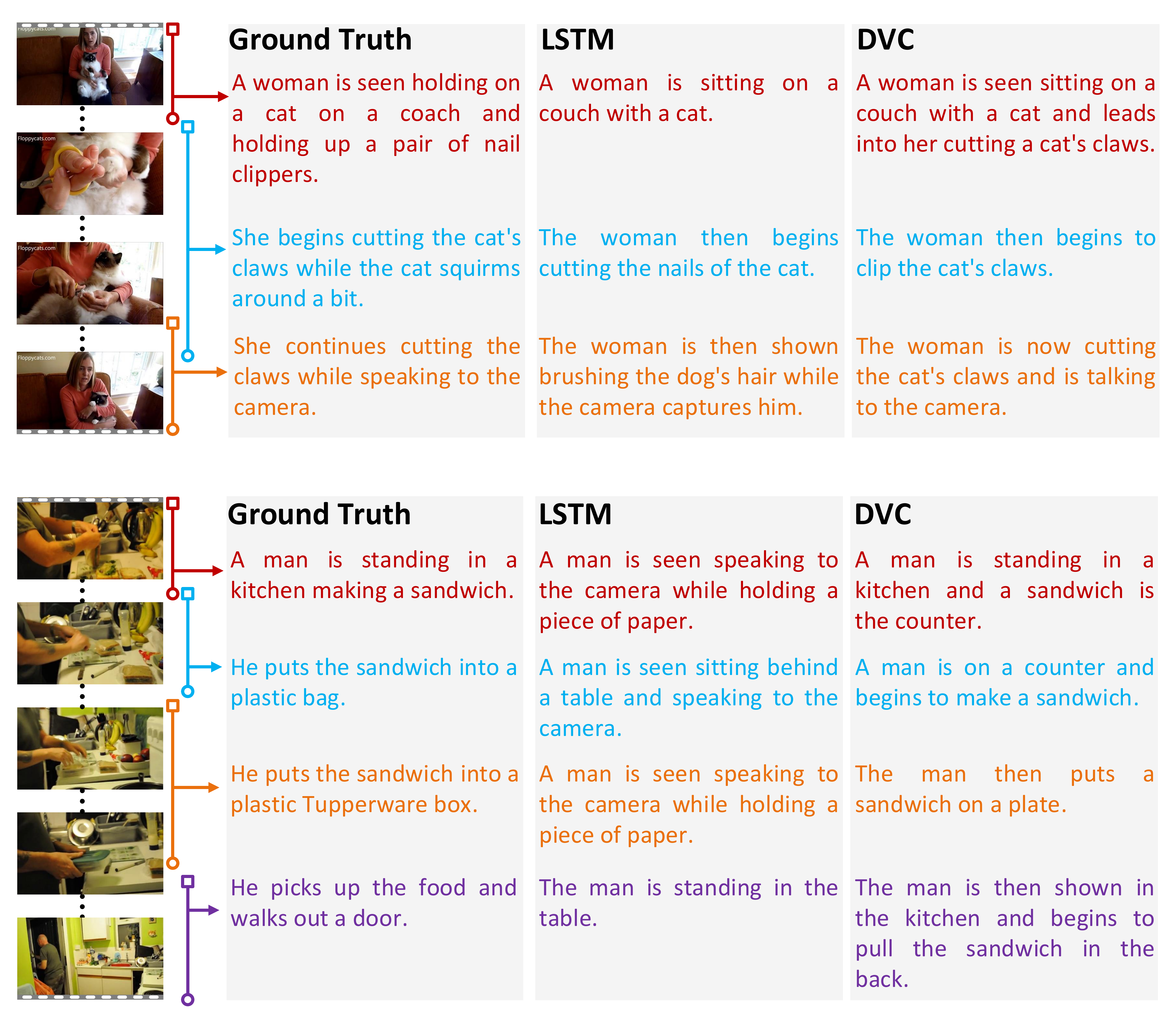}}
   \vspace{-0.1in}
   \caption{\small Dense video captioning results on ActivityNet Captions validation set. The output temporally localized sentences are generated by 1) Ground Truth, 2) LSTM, and 3) our DVC. We show the results with the highest overlap with ground truth captions.}
   \label{fig:figs_example}
   \vspace{-0.26in}
\end{figure}

\textbf{Performance Comparison.}
Table \ref{tab:densecaption} shows the performances of different models on ActivityNet Captions validation set. Overall, the results across six evaluation metrics with ground truth proposals and learnt proposals consistently indicate that our proposed DVC achieves superior performances against other state-of-the-art video captioning techniques including non-attention models (LSTM, S2VT) and attention-based approaches (TA, H-RNN, DCE). In particular, the METEOR score of our DVC can achieve 10.33\% with ground truth proposals, making the relative improvement over the best competitor DCE by 16.33\%, which is considered as a significant progress on this benchmark. As expected, the METEOR score is dropped down to 6.93\% when provided the predicted proposals from our TEP module instead of ground truth proposals. Moreover, DVC-D by additionally leveraging descriptiveness-driven temporal attention for dense video captioning, leads to a performance boost against LSTM. The result basically indicates the advantage of weighting each local clips with holistic attention score in producing proposal representation for sentence generation, instead of representing each proposal by directly performing ``mean pooling" over its clips in LSTM. In addition, DVC-D-A by additionally augmenting LSTM with high-level semantic attributes, consistently improves DVC-D over all the metrics, but the METEOR scores are still lower than DVC. This confirms the effectiveness of utilizing reinforcement learning techniques for directly optimizing LSTM with METEOR-based reward loss, which harmonizes SG module with respect to its testing inference.

Figure \ref{fig:figs_example} showcases a few dense video captioning results generated by different methods and human-annotated ground truth sentences. From these exemplar results, it is easy to see that all of these automatic methods can generate somewhat relevant sentences, while our proposed DVC can generate more relevant and descriptive sentences by jointly exploiting descriptiveness-driven temporal attention mechanism and high-level attributes for boosting dense video captioning. For example, compared to phrase ``brushing the dog's hair" in the sentence generated by LSTM, ``cutting
the cat's claws" in our DVC is more precise to describe the event proposal in the last proposal of the first video.

\begin{table}\scriptsize
    \centering
    \caption{\small User study on two criteria: M1 - percentage of sets of captions generated by different methods that are evaluated as better/equal to human captions; M2 - percentage of sets of captions that pass Turing Test.}
    \label{table:user}
    \setlength{\tabcolsep}{2pt}
    \begin{tabular}{c|c|c|c|c|c|c}
        \Xhline{2\arrayrulewidth}
        & ~Human~ & ~DVC~ & ~H-RNN \cite{Yu:CVPR16}~ & ~TA \cite{Yao:ICCV15}~ & ~S2VT \cite{Venugopalan:ICCV15}~ & ~LSTM \cite{Venugopalan:NAACL15}~ \\
        \hline
        ~~~M1~~~             & -         & \textbf{35.1}        & 32.9        & 30.8          & 30.2      & 28.7      \\
        ~~~M2~~~             & \textbf{97.6}     & \textbf{38.7}        & 36.3        & 34.6          & 34.1      & 31.9      \\
        \Xhline{2\arrayrulewidth}
    \end{tabular}
    \vspace{-0.25in}
\end{table}

\textbf{Human Evaluation.}
To better understand how satisfactory are the localized temporal event proposals and the corresponding generated sentences of different methods, we also conducted a human study to compare our DVC against four baselines, i.e., H-RNN, TA, S2VT, and LSTM. A total number of 12 evaluators from different education backgrounds are invited and a subset of 1K videos is randomly selected from validation set for the subjective evaluation. The evaluation process is as follows. All the evaluators are organized into two groups. We show the first group all the five sets of temporally localized sentences generated by each approach plus a series of temporally human-annotated sentences and ask them the question: Do the systems produce the set of temporally localized sentences resembling human-generated sentences? In contrast, we show the second group once only one set of temporally localized sentences generated by different approach or human annotations and they are asked: Can you determine whether the given set of sentences has been generated by a system or by a human being? From evaluators' responses, we calculate two metrics: 1) M1: percentage of sets of captions that are evaluated as better or equal to human captions; 2) M2: percentage of sets of captions that pass the Turing Test. Table \ref{table:user} lists the result of the user study. Overall, our DVC is clearly the winner for all two criteria. In particular, the percentage achieves 35.1\% and 38.7\% in terms of M1 and M2, respectively, making the absolute improvement over the best competitor H-RNN by 2.2\% and 2.4\%.

\begin{table}\scriptsize
    \centering
    \caption{\small Leaderboard of the published state-of-the-art dense video captioning models on the online ActivityNet evaluation server.}
    \label{table:leaderboard}
    \setlength{\tabcolsep}{2pt}
    \begin{tabular}{c|c|c|c|c}
        \Xhline{2\arrayrulewidth}
          & ~~~DVC (P3D)~~~ & ~~~DVC (C3D)~~~ & ~~~~~TAC \cite{ActivityNet}~~~~~ & ~~~~~DCE \cite{krishna2017dense}~~~~~ \\
        \hline
        ~~~METEOR~~~         & \textbf{12.96}            & \textbf{10.36}               & 9.61           & 4.82             \\
        \Xhline{2\arrayrulewidth}
    \end{tabular}
    \vspace{-0.23in}
\end{table}

\textbf{Performance on ActivityNet Evaluation Server.}
We also submitted our best run in terms of METEOR score, i.e., DVC, to online ActivityNet evaluation server and evaluated the performance on official testing set. Table \ref{table:leaderboard} shows the performance Leaderboard on official testing set. Please note that here we design two submission runs for our DVC, i.e., DVC (C3D) and DVC (P3D). The input clip features of the two runs are 500-way C3D features and 2048-way output of $pool5$ layer from P3D ResNet \cite{qiu2017learning}, respectively. Compared to the top performing methods, our proposed DVC (C3D) achieves the best METEOR score. In addition, when leveraging the clip feature from P3D ResNet, our METEOR score on testing set is further boosted up to 12.96\%, ranking the first on the Leaderboard.

\subsection{Temporal Event Proposal Task}
The second experiment is conducted on temporal event proposal task, which evaluates our TEP module's capability to adequately localize all events for a given video.

\textbf{Compared Approaches.}
We compare our DVC with three state-of-the-art temporal action proposal methods:

(1) Temporal Actionness Grouping (TAG) \cite{zhao2017temporal}. TAG utilizes actionness classifier to generate actionness curve, followed by the watershed algorithm to produce basins. The proposals are finally generated by grouping the basins.

(2) Dense-Captioning Events (DCE) \cite{krishna2017dense}. DCE leverages a multi-scale variant of LSTM-based action proposal model in \cite{escorcia2016daps} to localize temporal event proposals.

(3) Temporal Unit Regression Network (TURN) \cite{gao2017turn}. TURN jointly predicts action proposals and refines the temporal boundaries by temporal coordinate regression.

\textbf{Evaluation Metrics.}
For temporal action proposal task, we adopt the Area-Under-the-Curve (AUC) score for Average Recall vs. Average Number of Proposals per Video (AR-AN) curve in \cite{ActivityNet} as the evaluation metric. AR is defined as the mean of all recall values using tIoU thresholds between 0.5 and 0.95 (step size: 0.05), and AN denotes the total number of proposals divided by the number of videos.

\begin{figure}[!tb]
   \centering {\includegraphics[width=0.35\textwidth]{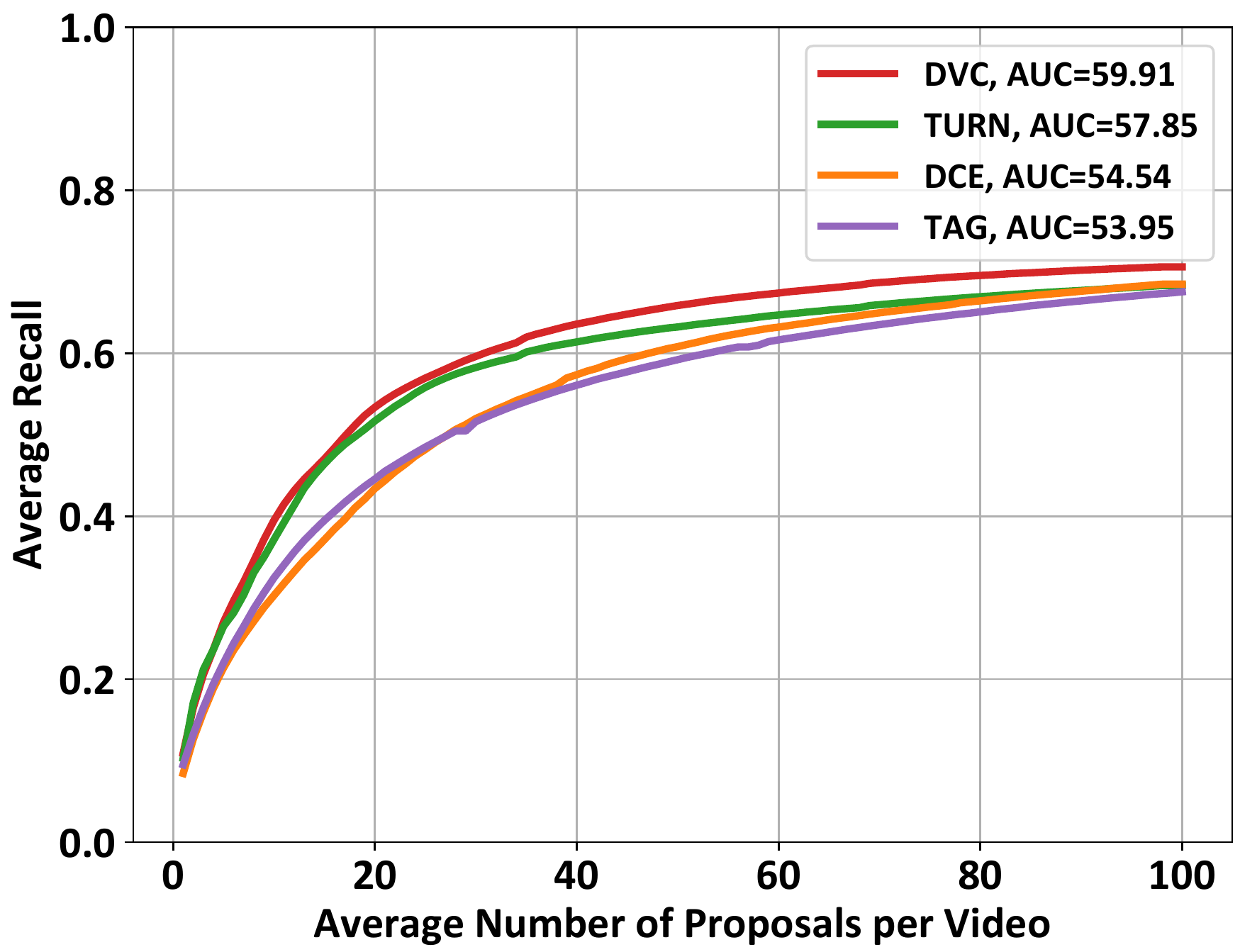}}
   \caption{\small The AR-AN curves of different approaches on ActivityNet Captions validation set for temporal event proposal task.}
   \label{fig:figTEP}
   \vspace{-0.12in}
\end{figure}

\textbf{Performance Comparison.}
Figure \ref{fig:figTEP} shows the AR-AN curves of four runs on ActivityNet Captions validation set for temporal event proposal task. Overall, the quantitative results with regard to AUC score indicate that our DVC outperforms other methods. In particular, by leveraging temporal coordinate regression for adjusting temporal boundary of detected proposals, DCE and TURN lead a large performance boost against TAG. Moreover, DVC by additionally incorporating descriptiveness regression into TEP module further improves DCE and TURN. The result indicates the advantage of joint detection and captioning, which refines the event proposal from language perspective.

\textbf{Effects of Multiple Anchor Layers.}
In order to show the relationship between performance and the number of anchor layers with different temporal resolutions, we progressively stack anchor layers with decreasing temporal resolutions and compare the performances. The results shown in Table \ref{tab:multilayers} indicate increasing the number of anchor layers with different temporal resolutions can generally lead to performance improvements. Meanwhile, the number of parameters in all adopted anchor layers increases. Thus, we finally adopt $conv_{3 \sim {11}}$ as anchor layers as that has a better tradeoff between performance and model complexity.

\begin{table}\scriptsize
    \centering
    \setlength{\tabcolsep}{2pt}
    \caption{Effects of utilizing multiple anchor layers.}
    \begin{tabular}{ccccc|cc}
        \Xhline{2\arrayrulewidth}
        $conv_{3 \sim {8}}$ & $conv_9$   & $conv_{10}$   & $conv_{11}$   & $conv_{12}$  & ~~AUC~~   & ~~Parameter Number~~     \\
        \hline \hline
        \ding{52}        &           &           &           &           & 55.73       &  16.5M   \\
        \ding{52}        & \ding{52} &           &           &           & 57.38       &  18.9M   \\
        \ding{52}        & \ding{52} & \ding{52} &           &           & 58.75       &  21.2M   \\
        \ding{52}        & \ding{52} & \ding{52} & \ding{52} &           & 59.91       &  23.6M \\
        \ding{52}        & \ding{52} & \ding{52} & \ding{52} & \ding{52} & 60.07       &  26.0M\\
        \Xhline{2\arrayrulewidth}
    \end{tabular}
    \label{tab:multilayers}
      \vspace{-0.2in}
\end{table}

\section{Conclusions}
We have presented a novel deep architecture which unifies the temporal localization of event proposals and sentence generation for dense video captioning. Particularly, we study the problems of how to build the interaction across the two sub challenges (i.e., temporal event proposal and sentence generation) and how to integrate such interaction into a deep learning framework for enhancing dense video captioning. To verify our claim, we have devised a descriptiveness regression component and incorporated it into a single shot detection structure, on one hand to adjust the event proposal from language perspective in TEP module, and on the other, to measure the descriptive complexity of each event in SG module. Experiments conducted on ActivityNet Captions dataset validate our model and analysis. More remarkably, we achieve superior results over state-of-the-art methods when evaluating our framework on both dense video captioning and temporal event proposal tasks.

\textbf{Acknowledgments.} This work was supported in part by NSFC under Grant 61672548, U1611461, and the Guangzhou Science and Technology Program, China, under Grant 201510010165.

{\small
\bibliographystyle{ieee}
\bibliography{egbib}
}

\end{document}